\documentclass[sigconf,authorversion]{acmart}

\AtBeginDocument{%
  \providecommand\BibTeX{{%
    \normalfont B\kern-0.5em{\scshape i\kern-0.25em b}\kern-0.8em\TeX}}}

\usepackage{amsmath,amsfonts}
\usepackage{algorithm}
\usepackage[noend]{algpseudocode}
\usepackage{graphicx}
\usepackage{textcomp}
\usepackage{xcolor}
\usepackage{svg}
\usepackage{comment}
\usepackage{placeins}
\usepackage{balance}
\usepackage{adjustbox}

\usepackage[skip=0.33\baselineskip]{caption}
\usepackage{subcaption}
\setcopyright{acmcopyright}
\copyrightyear{2024}
\acmYear{2024}
\setcopyright{rightsretained}
\acmConference[ICMLSC 2024]{2024 The 8th International Conference on Machine Learning and Soft Computing}{January 26--28, 2024}{Singapore, Singapore}
\acmBooktitle{2024 The 8th International Conference on Machine Learning and Soft Computing (ICMLSC 2024), January 26--28, 2024, Singapore, Singapore}\acmDOI{10.1145/3647750.3647762}
\acmISBN{979-8-4007-1654-6/24/01}

\begin{document}

\title{TensAIR: Real-Time Training of Neural Networks from Data-streams}

\author{Mauro D. L. Tosi}
\email{mauro.dalleluccatosi@uni.lu}
\orcid{0000-0002-0218-2413}
\affiliation{%
  \institution{University of Luxembourg}
  \country{Luxembourg}
}

\author{Vinu E. Venugopal}
\email{vinu.ev@iiitb.ac.in}
\orcid{0000-0003-4429-9932}
\affiliation{%
  \institution{IIIT Bangalore}
  \country{India}
}
\author{Martin Theobald}
\email{martin.theobald@uni.lu}
\orcid{0000-0003-4067-7609}
\affiliation{%
  \institution{University of Luxembourg}
  \country{Luxembourg}
}

\renewcommand{\shortauthors}{Tosi, et al.}

\begin{abstract}
Online learning (OL) from data streams is an emerging area of research that encompasses numerous challenges from stream processing, machine learning, and networking. Stream-processing platforms, such as Apache Kafka and Flink, have basic extensions for the training of Artificial Neural Networks (ANNs) in a stream-processing pipeline. However, these extensions were not designed to train ANNs in real-time, and they suffer from performance and scalability issues when doing so. 

This paper presents TensAIR, the first OL system for training ANNs in real time. TensAIR achieves remarkable performance and scalability by using a decentralized and asynchronous architecture to train ANN models (either freshly initialized or pre-trained) via DASGD (decentralized and asynchronous stochastic gradient descent). We empirically demonstrate that TensAIR achieves a {\em nearly linear scale-out} performance in terms of (1) the number of worker nodes deployed in the network, and (2) the throughput at which the data batches arrive at the dataflow operators. We depict the versatility of TensAIR by investigating both sparse (word embedding) and dense (image classification) use cases, for which TensAIR achieved from 6 to 116 times higher sustainable throughput rates than state-of-the-art systems for training ANN in a stream-processing pipeline. 
\end{abstract}



\maketitle

\section{Introduction}

Online learning (OL) is a branch of Machine Learning (ML) which studies solutions to time-sensitive problems that demand real-time answers based on fractions of data received in the form of {\em data streams} \cite{hoi2021online}. A common characteristic of data streams is the presence of {\em concept drifts} \cite{iwashita2018overview}, i.e., changes in the statistical properties among the incoming data objects over time \cite{lu2018learning}. Consequently, pre-trained ML models tend to be inadequate in OL scenarios as their performance usually decreases after concept drifts \cite{lu2018learning}. Differently, OL models mitigate the negative effects of such concept drifts by being ready to instantly update themselves for any new data from the data stream \cite{hoi2021online}.

Due to the intrinsic time-sensitiveness of OL, it is not feasible to depend on solutions that spend an undue amount of time on retraining. Thus, if concept drifts are frequent, currently, complex OL problems cannot rely on robust solutions common to other ML problems due to their long training time \cite{hoi2021online}, like those involving Artificial Neural Networks (ANNs) \cite{Goodfellow-et-al-2016}. 

Therefore, how to solve complex OL problems like those involving data streams of audio, video, or even text remains an open research question, especially when they are affected by frequent concept drifts. Currently, most OL researchers are focused on how to improve the quality of the input data and how to adapt to concept drifts \cite{lu2018learning,hu2020no,priya2021deep,barros2018large,sethi2017reliable} and they do not address the issue of solving such complex problems. Our intuition is that current hardware is already capable of training a large class of ANN models in real time if the training is distributed across multiple nodes efficiently. In this case, problems that today are deemed too complex to be effectively solved by standard OL learners could be solved using ANNs.

Nowadays, instead of retraining ANNs in real-time, state-of-the-art extensions \cite{dl-on-flink, kafka_tensorflow} for Apache Flink \cite{carbone2015apache} and Kafka \cite{kreps2011kafka} were developed to adapt/re-train their models using datasets created from buffered samples from the data-stream. Thus, these approaches enable the usage of ANN in OL by giving up the real-time adaptation of the ANN models, which can only be updated after buffering a dataset substantially large to be used for retraining. If adapted to be trained in real-time, those approaches suffer from performance and scalability issues (cf. Section \ref{sec:experiments}). Thus, they cannot sustain throughput high enough for many real-world problems.

Consequently, when real-time adaptation is not available, one can expect a lower prediction/inference performance of models between the instant a {\em concept drift} occurs and the moment the model is updated. Thus, considering that non-trivial ANN models demand a high amount of training examples before convergence, one can expect low-quality predictions/inferences for an extended amount of time (until the training dataset is buffered and the model is retrained). This makes it unfeasible to apply this approach to real-world problems that suffer from frequent concept drifts. 

To mitigate the prediction/inference performance decrease, we argue that it is necessary to adapt the ANN models in real-time. However, the real-time adaptation of ANN models on an OL scenario is not straightforward. We therefore highlight the following two challenges:

\begin{itemize}
\item[(1)] {\em real-time data management:} Not all training data is available from the beginning. Thus, it is necessary to incrementally update the model with fractions of data at each step. Different from commonly used pre-defined training datasets.
\item[(2)] {\em backpressure:} The model must process a higher number of data samples per second (for training and for inference/ prediction) than the data stream produces. This avoids a sudden surge in latency or even a system crash.
\end{itemize}

In this paper, we present the architecture of TensAIR, the first OL framework for training ANN models (either freshly initialized or pre-trained) in real time. TensAIR leverages the fact that stochastic gradient descent (SGD) is an iterative method that can update a model based only on a fraction of the training data per iteration. Thus, instead of using pre-defined or buffered datasets for training, TensAIR models are updated after each data sample (or data batch) is made available by the data stream. In addition, TensAIR achieves remarkable scale-out performance by using a fully decentralized and asynchronous architecture throughout its whole dataflow, thus leveraging the usage of DASGD (decentralized and asynchronous SGD) to update the ANN models.

To assess TensAIR, we performed experiments on sparse (word embedding) and dense (image classification) models. In our experiments, TensAIR achieved nearly linear scale-out performance in terms of (1) the number of worker nodes deployed in the network, and (2) the throughput at which the data batches arrive at the dataflow operators. Moreover, we observed the same convergence rate in the distributed models independently of the number of worker nodes, which shows that the usage of DASGD did not negatively impact the models' convergence. When compared to the state of the art, TensAIR's sustainable throughput in the real-time OL setting was from 6 to 175 times higher than Apache Kafka extension \cite{kafka_tensorflow} and from 6 to 120 times higher than Apache Flink extension \cite{dl-on-flink}. We additionally compared TensAIR to Horovod \cite{sergeev2018horovod}, distributed ANN framework developed by Uber, and achieved from 4 to 335 times higher sustainable throughput than them in the same real-time OL setting.

Below, we summarize the main contributions of this paper.

\medskip\noindent\textbf{Contributions~}
\begin{enumerate}
    \item Design and implementation of TensAIR, the first framework for real-time training and prediction in ANN models. 
    \item Creation and usage of our Decentralized and Asynchronous SGD (DASGD) algorithm. 
    \item Experimental evaluation of TensAIR showing almost linear training time speed-up in terms of nodes deployed.
    \item Sustainable throughput comparison between TensAIR and state-of-the-art systems, with TensAIR achieving from 4 to 120 times higher sustainable throughput than the baselines;
    \item Depiction of real-time Sentiment Analysis use case that would not be feasible with standard OL approaches. 
\end{enumerate}

\section{Background}
\label{sec:background}

Considering that the real-time training of ANNs in an OL scenario involves multiple areas of research, we give in the following subsections a short summary of the most important concepts and techniques used in this paper. 

\subsection{Online Learning}
Online learning (OL) has gained visibility due to the increase in the velocity and volume of available data sources compared to the past decade \cite{gomes2019machine}. OL algorithms are trained using data streams as input, which differs from traditional ML algorithms that have a pre-defined training dataset.

\smallskip\noindent\textbf{Streams \& Batches.~} Formally, a data stream $\mathcal{S}$ consists of ordered {\em events} $e$ with {\em timestamps} $s$, i.e., $(e_1,s_1), \ldots, (e_\infty,s_\infty)$, where the $s_i$ denote the processing time at which the corresponding events $e_i$ are ingested into the system. These events are usually analysed in {\em batches} $B_j$ of fixed size $b$, as follows:
\begin{align}
B_1 &= (e_1,s_1), \ldots, (e_b,s_b)\nonumber\\
B_2 &= (e_{b+1},s_{b+1}), \ldots, (e_{2b},s_{2b})\nonumber\\
&\ldots\nonumber
\end{align}

Batches $B_j$ are analyzed individually. Thus, if processed in an asynchronous stream-processing scenario, the batches (and in particular the included events $e_i$) can become out-of-order as they are handled within the system, even if the initial $s_i$ were ordered. In common stream-processing architectures, such as Apache Flink \cite{carbone2015apache}, Spark \cite{zaharia2016apache} and Samza \cite{noghabi2017samza}, batches are distinguished into {\em sliding windows}, {\em tumbling windows} and (per-user) {\em sessions} \cite{10.14778/2824032.2824076}.

\smallskip\noindent\textbf{Latency vs. Throughput.~} When analyzing systems that process data streams, one typically benchmarks them by their latency and throughput \cite{karimov2018benchmarking}. Formally, {\em latency} is the time it takes for a system to process an event, from the moment it is ingested to the moment it is used to produce a desired output. {\em Throughput}, on the other hand, is the number of events that a system can receive and process per time unit. The {\em sustainable throughput} is the maximum throughput at which a system can handle a stream over a sustained period of time (i.e., without exhibiting a sudden surge in latency, then called ``backpressure'' \cite{10.1145/2723372.2742788}, or even a crash).

\smallskip\noindent\textbf{Passive \& Active Drift Adaptation.~} To adapt to concept drifts, one may rely on either passive or active adaptation strategies \cite{heusinger2020passive}. The passive strategy updates the trained model indefinitely, with no regard to the actual presence of concept drifts. Active drift adaptation strategies, on the other hand, only adapt the model when a concept drift has been explicitly identified.

\subsection{Artificial Neural Networks}
ANNs denote a family of supervised ML algorithms which are designed to be trained on a pre-defined dataset \cite{Goodfellow-et-al-2016}. A training dataset is composed of multiple $(x,y)$ pairs, in which $x$ is a training example and $y$ is its corresponding label. ANNs are usually trained using {\em mini-batches} $X$, which are sets of $(x,y)$ pairs of fixed size $N$ that are iteratively (randomly) sampled from the training dataset, thus $X = {(x_i,y_i),\ldots,(x_{i+N},y_{i+N})}$. 

An ANN model is represented by the weights and biases of the network, described together by $\theta$ and it is usually trained with variants of {\em stochastic gradient descent} (SGD) \cite{robbins1951stochastic}. SGD updates $\theta$ by considering $\nabla L(X,\theta)$, which is the gradient of a pre-defined {\em loss function} $L$ with respect to $\theta$ when taking $X$ as input. Thus, we can represent the update rule of $\theta$ as in Equation \ref{eq:gradient_descent}, in which $t$ is the iteration in SGD, and $\alpha$ is a pre-defined learning rate.
\begin{equation}
\theta_{t+1} = \theta_t - \alpha \nabla L(X,\theta_t)
\label{eq:gradient_descent}
\end{equation}

Based on Equation \ref{eq:gradient_descent}, $\theta_{t+1}$ is defined based on two terms. The second term is the more computationally expensive one to calculate, which we refer to as \textit{gradient calculation} (GC). The remainder of the equation we call \textit{gradient application} (GA), which consists of the subtraction between the two terms and the assignment of the result to $\theta_{t+1}$.    

\smallskip\noindent\textbf{Distributed Artificial Neural Networks.~}
Over the last years, ANN models have substantially grown in size and complexity. Consequently, the usage of traditional centralized architectures has become unfeasible when training complex models due to the high amount of time they spend until convergence \cite{sergeev2018horovod}. Researchers have been studying how to distribute ANN training to mitigate this.
Distributed ANNs reduce the time it takes to train a complex ANN model by distributing its computation across multiple compute nodes. This distribution can follow different parallelization methods, system architectures, and synchronisation settings \cite{mayer2020scalable}.

The most common form of distributing ANNs, which we also use in this work, is referred to as {\em data parallelism} \cite{ouyang2021communication}, in which workers are initialized with replicas of the same initial model and trained with disjoint splits of the training data. Moreover, the synchronisation among the workers' parameters in a data-parallel ANN setting is either {\em centralised} or {\em decentralised} \cite{mayer2020scalable}. In a centralised architecture \cite{ouyang2021communication}, workers systematically send their parameter updates to one or multiple parameter servers. Those servers aggregate the updates of all workers and apply them to a centralised model \cite{mayer2020scalable}. Thus, by relying on parameter servers to aggregate updates, the parameter servers may become the bottleneck of such an architecture \cite{chen2019round}. On the other hand, in a decentralised architecture \cite{ouyang2021communication}, the workers synchronize themselves using a broadcasting-like form of communication \cite{ouyang2021communication}. This broadcast eliminates the bottleneck of the parameter servers but requires a direct communication among worker nodes. 

The parameter updates in a data-parallel ANN system can be {\em synchronous} or {\em asynchronous}. In a synchronous setting \cite{ouyang2021communication}, workers have to synchronize themselves after each mini-batch iteration. This synchronization barrier wastes computational resources at idle times (i.e., when workers have to wait for others to resume their computation) \cite{mayer2020scalable}. In an asynchronous SGD (ASGD) setting \cite{ouyang2021communication}, workers are allowed to compute their gradient computations also on stale model parameters. This behaviour obviously minimizes idle times but makes it harder to mathematically prove SGD convergence. Recent developments on ASGD \cite{jiang2017heterogeneity,sun2017asynchronous,zhang2020taming}, however, have tackled exactly this issue under different assumptions. Zhang et al. \cite{zhang2020taming} recently proved an $\mathbf{o}(1/\sqrt{k})$ convergence rate for unbounded non-convex problems using ASGD under a centralised parameter server setup (where $k$ denotes the iteration among the ASGD updates). Additionally, \cite{lian2018asynchronous,DBLP:conf/iclr/BornsteinRWBH23,tosi2023convergence} proved the convergence of ASGD on decentralized networks under distinct assumptions and network topologies. 
\section{TensAIR}
\label{sec:TensAIR}

We now introduce the architecture of {\em TensAIR}, the first framework for training and predicting in ANNs models in real-time. TensAIR was designed to work in association with stream-processing engines that allow asynchronous and decentralized 
communication among dataflow operators.

\begin{figure}[ht]
  \centering
  \includegraphics[width=0.5\textwidth]{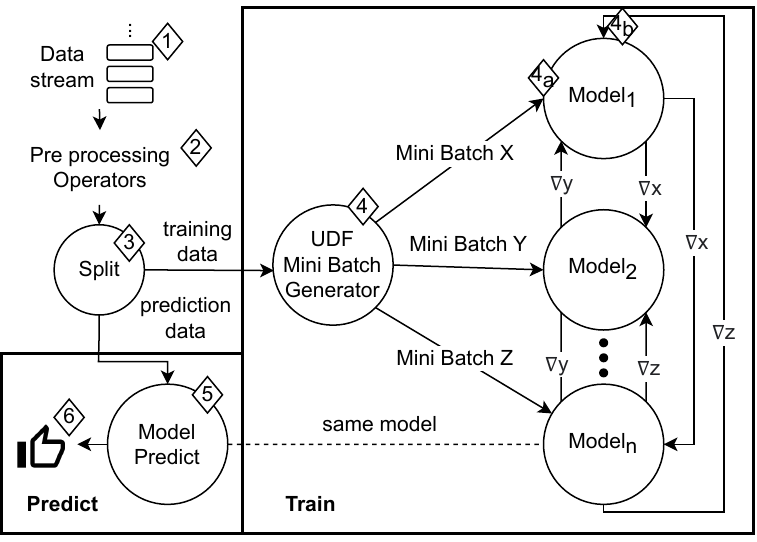}
  \caption{TensAIR generic dataflow with $n$ distributed {\tt Model} instances.}
  \label{fig:dataflow}
\end{figure}

TensAIR introduces the {\em data-parallel}, {\em decentralized}, {\em asynchronous} ANN operator {\tt Model}, with {\tt train} and {\tt predict} as two new OL functions. This means that TensAIR can scale out both the training and prediction tasks of an ANN model to multiple compute nodes, either with or without GPUs associated with them. TensAIR dataflow can be visualized using a graph (see Figure~\ref{fig:dataflow}). Note that, throughout this paper, we use the terms {\tt prediction} and {\tt inference} interchangeably. 

\medskip\noindent\textbf{TensAIR Dataflows.~} Figure~\ref{fig:dataflow} depicts a generic {\em TensAIR dataflow}. This dataflow is composed of a single input data stream, \textit{n} instances of the ${\tt Model}$ operator, and single instances of the {\tt Split} and {\tt UDF} operators. The idea behind a TensAIR dataflow is: (1) to receive training samples from the input data streams; (2) to pre-process the data received using common dataflow operators like {\tt Map}, {\tt Reduce}, {\tt Split}, and {\tt Join} to transform the data as deemed necessary given each use case; (3) to select whether the pre-processed data samples will be used for training, for prediction, or for both; (4) if data is sent for training, to aggregate a pre-defined number of samples in the form of a mini-batch by using a {\tt user-defined function UDF}, and to send this mini-batch to one of the decentralized {\tt Model} instances; (4a) when a {\tt Model} instance receives a mini-batch $X$ from the {\tt UDF}, to calculate an update $\nabla x$ based on the current {\tt Model} weights and the mini-batch $X$, to apply the update to itself, and to broadcast the update to other {\tt Model} instances; (4b) when $Model_i$ receives an update from $Model_j$, to apply the received update locally; (5) if the pre-processed data is sent for prediction, to randomly select one of the distributed models and use it to perform the prediction; (6) to use the prediction previously made as final output of the dataflow or as input of further operators (as deemed necessary by the given use case).

\medskip\noindent\textbf{Stream Processing.~}
 As shown in Algorithm \ref{alg:model}, a TensAIR {\tt Model} operator has two new OL functions {\tt train} and {\tt predict}, which can asynchronously send and receive messages to and from other operators. During {\tt train}, {\tt Model} receives either {\em encoded mini-batches} $X$ or {\em gradients} $\nabla x$ as messages. Each message encoding a gradient that was computed by another model instance is immediately used to update the local model accordingly. Each mini-batch first invokes a local gradient computation and is then used to update the local model. Each such resulting gradient is also locally summed until a desired number of gradients ($\mathit{maxGradBuffer}$) is reached, upon which the buffer then is broadcast to all other {\tt Model} instances.

 \begin{algorithm}[H]
     \centering
     \caption{TensAIR {\tt Model} class}
     \begin{algorithmic}[1]
     \State \textbf{Constructor} {Model} ({$\mathit{tfModel}$, $\mathit{maxBuffer}$}):
        \State \hskip1.5em model = $\mathit{tfModel}$
        \State \hskip1.5em maxGradBuffer = $\mathit{maxBuffer}$
        \State \hskip1.5em gradients = $\emptyset$
        \State \hskip1.5em gradients\_count = 0
        
    \Procedure{process\_msg}{$\mathit{msg}$}
            \If{$\mathit{msg}.\mathit{mode}$ == TRAIN}
                \State {\tt train}($\mathit{msg}$)
            \Else
                \State {\tt predict}($\mathit{msg}$)
            \EndIf
    \EndProcedure
    
    \Procedure{train}{$\mathit{msg}$}
        \If{$\mathit{msg}.\mathit{isGradient}$}
            \State model = apply\_gradient(model, $\mathit{msg}$)
        \Else
            \State gradient = calculate\_gradient(model, $\mathit{msg}$)
            \State model = apply\_gradient(model, gradient)
            \State gradients += gradient
            \State gradients\_count += 1
            \If{gradients\_count $\geq$ maxGradBuffer}
                \State send\_gradients(gradients)
                \State gradients\_count = 0
            \EndIf
        \EndIf
    \EndProcedure

    \Procedure{predict}{$\mathit{msg}$}
        \State predictions = model.make\_predictions($\mathit{msg}$)
        \State send\_results(predictions)
    \EndProcedure
     \end{algorithmic}
     \label{alg:model}
 \end{algorithm}

\subsection{Model Consistency}
\label{sec:consistency}
Despite TensAIR's asynchronous nature, it is necessary to maintain the models consistent among themselves during training in order to guarantee that they are aligned and, therefore, they eventually convergence to a same common model. In TensAIR, this is given by the exchange of gradients between the various {\tt Model} instances. 

\begin{figure}[htb]
    \centering
    \includegraphics[width=0.75\linewidth]{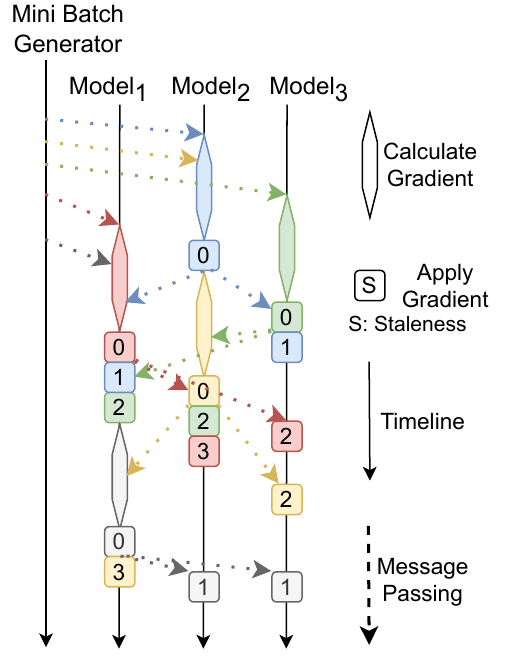}
        \caption{DASGD staleness calculation.}
        \label{fig:sync}
\end{figure}

Due to our asynchronous computation and application of the gradients on the distributed model instances, {\tt Model}$_i$ receives gradients calculated by {\tt Model}$_j$ (with $j \neq i$) which are similar but not necessarily equal to itself. This occurs whenever {\tt Model}$_i$, which has already applied to itself a set of $G_i=\{\nabla x, \nabla y, ..., \nabla z\}$ gradients, calculates a new gradient $\nabla a$, and sends it to {\tt Model}$_j$, such that $G_i \neq G_j$ at the time when {\tt Model}$_j$ applies $\nabla a$. The difference $|G_i \cup G_j|-|G_i \cap G_j|$ between these two models is defined as \textit{staleness} \cite{tosi2022convergence}. This $\mathit{staleness}_{i,j}(\nabla_{a})$ metric is the symmetric distance between $G_i$ and $G_j$ with respect to the times at which a new gradient $\nabla_a$ was computed by a model $i$ and is applied to model $j$, respectively. 
We illustrate this phenomenon and the staleness metric in Figure \ref{fig:sync}.

Figure \ref{fig:sync} illustrates the timeline of messages (containing both mini-batches and gradients) exchanged among TensAIR models considering $\mathit{max}\-\mathit{GradBuffer} = 1$. Assume the {\tt UDF} distributes 5 mini-batches to 3 models. After receiving their first mini-batch, each {\tt Model}$_i$ calculates a corresponding gradient. Note that, when applied locally, the staleness of any gradient is 0 because it is computed and immediately applied by the same model. While computing or applying a local gradient, each {\tt Model}$_i$ may receive more gradients to calculate and/or apply from either the {\tt UDF} or other models {\em asynchronously}. In our protocol, the models first finish their current gradient computation, apply it locally, then buffer and send $\mathit{maxGradBuffer}$ many locally computed gradients to the other models, and wait for their next update. 

As an illustration, take a look at {\tt Model}$_2$ in Figure \ref{fig:sync}. While computing $\nabla_{blue}$, it receives the yellow mini-batch from the {\tt Mini Batch generator}, which it starts computing immediately after it finishes processing the blue one---which it had already started when it received the yellow mini-batch. During the computation of $\nabla_{yellow}$, {\tt Model}$_2$ receives $\nabla_{green}$ to apply, which it does promptly after finishing $\nabla_{yellow}$. Note that when {\tt Model}$_3$ computed $\nabla_{green}$ and {\tt Model}$_1$ computed $\nabla_{red}$, they have not applied a single gradient to their local models at that time. Thus, $|G_{1}| = |G_{3}| = 0$. However, before applying $\nabla_{green}$, $G_{2} = \{\nabla_{blue}, \nabla_{yellow}\}$ with $|G_{2}| = 2$ and $\mathit{staleness}_{3,2}(\nabla_{green}) = 2$. Along the same lines, before applying $\nabla_{red}$, $|G_{2}| = 3$ and $\mathit{staleness}_{1,2}(\nabla_{red})= 3$.

\subsection{Model Convergence}
\label{sec:convergence}
Since TensAIR operates on data streams and is both asynchronous and fully decentralized (i.e., it has no centralized parameter server), it exhibits characteristics that most SGD proofs of convergence \cite{zhang2020taming,sun2017asynchronous,jiang2017heterogeneity} do not cover. Therefore, we next discuss under which circumstances TensAIR is guaranteed to converge.

First, we consider that training is performed between significant concept drifts. Therefore, we assume that the data distribution between two subsequent concept drifts does not change. Thus, if a concept drift occurs during the training, the model will not converge until the concept drift ends. By considering this, the data stream between two concept drifts will behave like a fixed data set. In this case, if given enough training examples, as seen in \cite{Goodfellow-et-al-2016}, each of the local model instances will eventually converge. 

Second, considering TensAIR's decentralized and asynchronous SGD (DASGD), model updates can be staled. Nevertheless, as proven by Tosi and Theobald \cite{tosi2023convergence}, the model will converge in this setting in up to $\mathcal{O}(\frac{\sigma}{\epsilon^2}) + \mathcal{O}(\frac{QS_{avg}}{\epsilon^\frac{3}{2}}) + \mathcal{O}(\frac{S_{avg}}{\epsilon})$ iterations to an $\epsilon$-small error, considering $S_{avg}$ as the average staleness observed during training and $Q$ a constant that bounds the gradients size. If bounded gradients are not assumed, DASGD converges in $\mathcal{O}(\frac{\sigma}{\epsilon^2}) + \mathcal{O}(\frac{\sqrt{\hat{S}_{avg}\hat{S}_{max}}}{\epsilon})$ iterations, with $\hat{S}_{max}$ and $\hat{S}_{avg}$ representing the maximum and average staleness, calculated using an additional recursive factor. 

\subsection{Implementation}

TensAIR was implemented on top of the Asynchronous Iterative Routing (AIR) \cite{venugopal2020air, venugopal2022targeting} dataflow engine. AIR is a native stream-processing engine that processes complex dataflows in an asynchronous and decentralized manner. {\em TensAIR dataflow} operators extend a basic {\tt Vertex} superclass in AIR. {\tt Vertex} implements AIR's asynchronous MPI protocol via multi-threaded queues of incoming and outgoing messages, which are exchanged among all nodes (aka. ``ranks'') in the network asynchronously. This is crucial to guarantee that worker nodes do not stay idle while waiting to send or receive messages during training. The number of instances of each {\tt Vertex} subclass and the number of input data streams can be configured beforehand, as seen in Figure \ref{fig:dataflow}.

TensAIR is completely implemented in C++. It includes the TensorFlow 2.8 native C API to load, save, train, and predict ANN models. Therefore, it is possible to develop a TensorFlow/Keras model in Python, save the model to a file, and load it directly into TensAIR. TensAIR is completely open-source and available from our GitHub repository\footnote{\url{https://github.com/maurodlt/TensAIR}}.

\section{Experiments \& Discussion}
\label{sec:experiments}

To assess TensAIR, we performed experiments to measure its performance on solving prototypical ML problems such as Word2Vec ({\em word embeddings}) and CIFAR-10 (image classification). We empirically validate TensAIR's model convergence by comparing its training loss curve at increasing levels of distribution across both CPUs and GPUs. Our results confirm that TensAIR's DASGD updates achieve similar convergence on Word2Vec and CIFAR-10 as a synchronous SGD propagation. At the same time, we achieve a nearly linear reduction in training time on both problems. Due to this reduction, TensAIR significantly outperforms not just the current OL extensions of Apache Kafka and Flink (based on both the standard and distributed TensorFlow APIs), but also Horovod which is a long-standing effort to scale-out ANN training. Finally, by providing an in-depth analysis of a \textit{sentiment analysis} (SA) use-case on Twitter, we demonstrate the importance of OL in the presence of concept drifts (i.e., COVID-19 related tweets with changing sentiments). In particular the SA usecase is an example of task that would be deemed too complex to be adapted in real-time (at a throughput rate of up to 6,000 tweets per second) when using other OL frameworks.

\smallskip\noindent\textbf{HPC Setup.~} We carried out the experiments described in this section using the HPC facilities of the University of Luxembourg \cite{VCPKVO_HPCCT22}. We distributed the ANNs training using up to 4 Nvidia Tesla V100 GPUs in a node with 768 GB RAM. We also deployed up to 16 regular nodes, with 28 CPU cores and 128 GB RAM each, for the CPU-based (i.e., without using GPU acceleration) settings.

\smallskip\noindent\textbf{Event Generation.~} We trained both sparse (word embeddings\footnote{https://www.tensorflow.org/tutorials/text/word2vec}) and dense (image classification\footnote{https://www.tensorflow.org/tutorials/images/cnn}) models based on English Wikipedia articles and images from CIFAR-10 \cite{krizhevsky2009learning}, respectively. Instead of connecting to actual streams, we chose those static datasets to facilitate a consistent analysis of the results and ensure reproducibility. Moreover, to simulate a streaming scenario, we implemented the \texttt{MiniBatchGenerator} as an entry-point \texttt{Vertex} operator (compare to Figure \ref{fig:dataflow}) which generates events $e_i$ with timestamps $s_i$, groups them into mini-batches $X_j$ by using a tumbling-window semantics, and sends these mini-batches to the subsequent operators in the dataflow. Furthermore, this allows us to simulate streams of unbounded size by iterating over the datasets multiple times (in analogy to training with multiple epochs over a fixed dataset).

\smallskip\noindent\textbf{Sparse vs. Dense Models.~} We chose Word2Vec and CIFAR-10 because they represent prototypical ML problems with {\em sparse} and {\em dense} model updates, respectively. Sparse updates mean that only a small portion of the neural network variables actually become updated per mini-batch \cite{recht2011hogwild}. Hence, sparseness should assist the models' convergence when using DASGD, as observed also in Hogwild! \cite{recht2011hogwild}. We trained by sampling 1\% from English Wikipedia which corresponds to 11.7M training examples (i.e., word pairs). On the other hand, we chose CIFAR-10 for being dense. Thus, we could analyze how this characteristic possibly hinders convergence when models are distributed and updated asynchronously. We train on all of the 50,000 labeled images of the CIFAR-10 dataset.

\subsection{Convergence Analysis} \label{sec:converge}
\begin{figure*}[ht]
    \centering
    \begin{subfigure}{0.42\linewidth}
        \centering
        \includegraphics[width=1\linewidth]{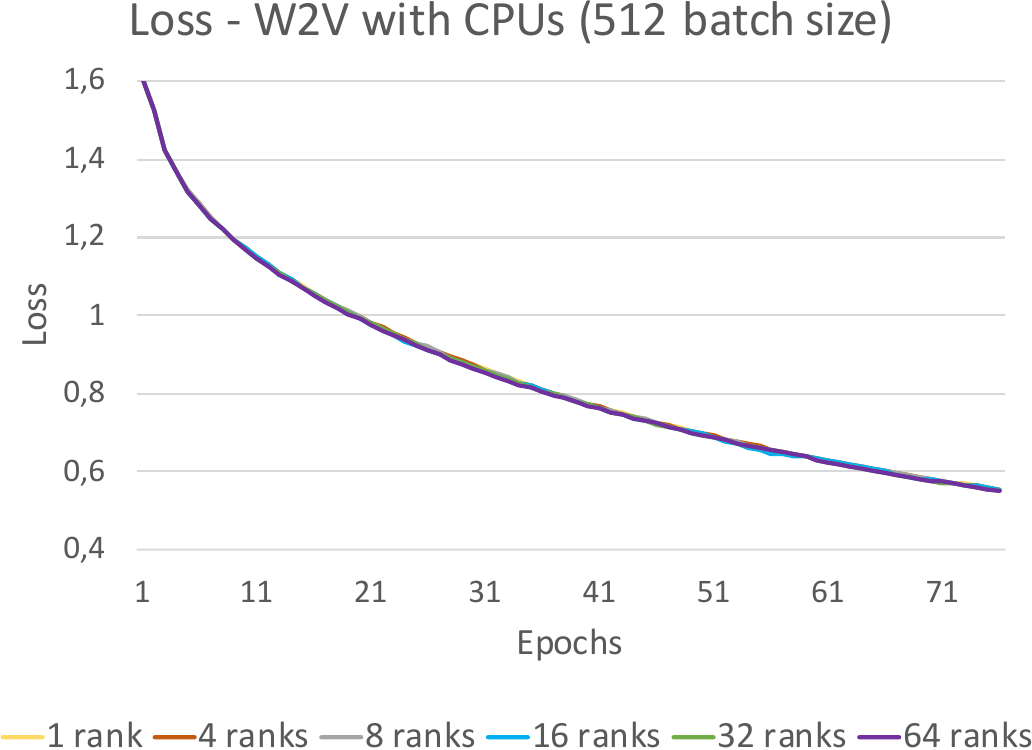}
        \caption{}
        \label{fig:Loss-W2V-CPU}
    \end{subfigure}%
    \hspace{0.07\textwidth}
    \begin{subfigure}{0.38\textwidth}
        \centering
        \includegraphics[width=1\linewidth]{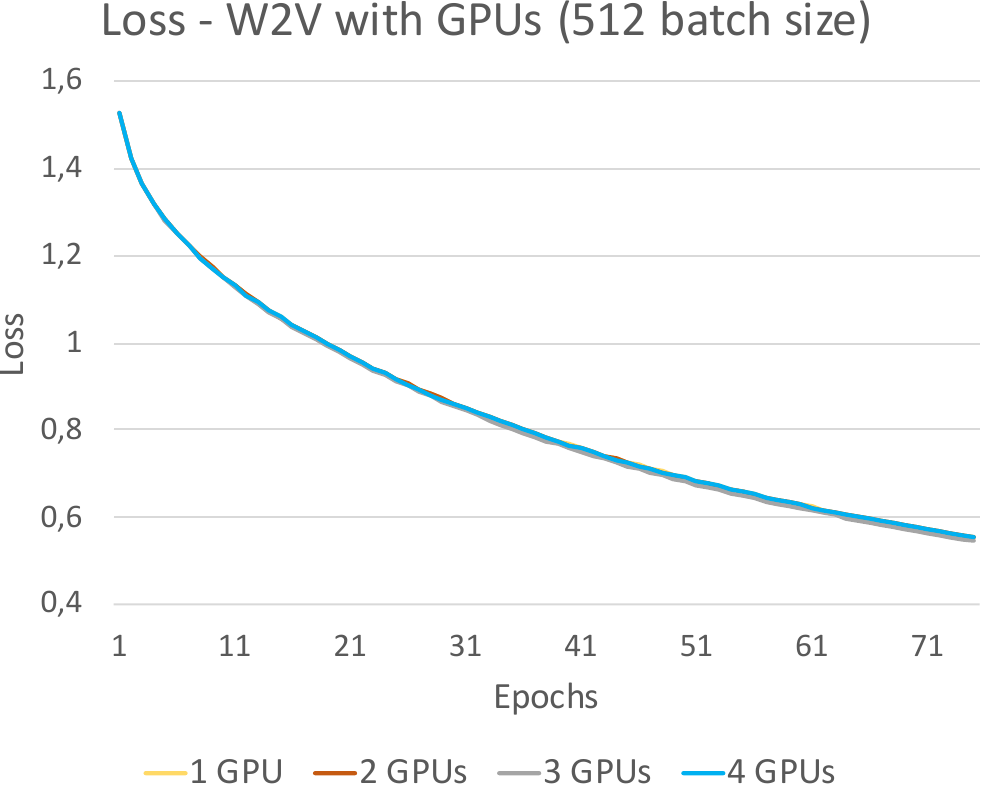}
        \caption{}
        \label{fig:Loss-W2V-GPU}
    \end{subfigure}%
    \vspace{1em}
    \begin{subfigure}{0.42\textwidth}
        \centering
        \includegraphics[width=1\linewidth]{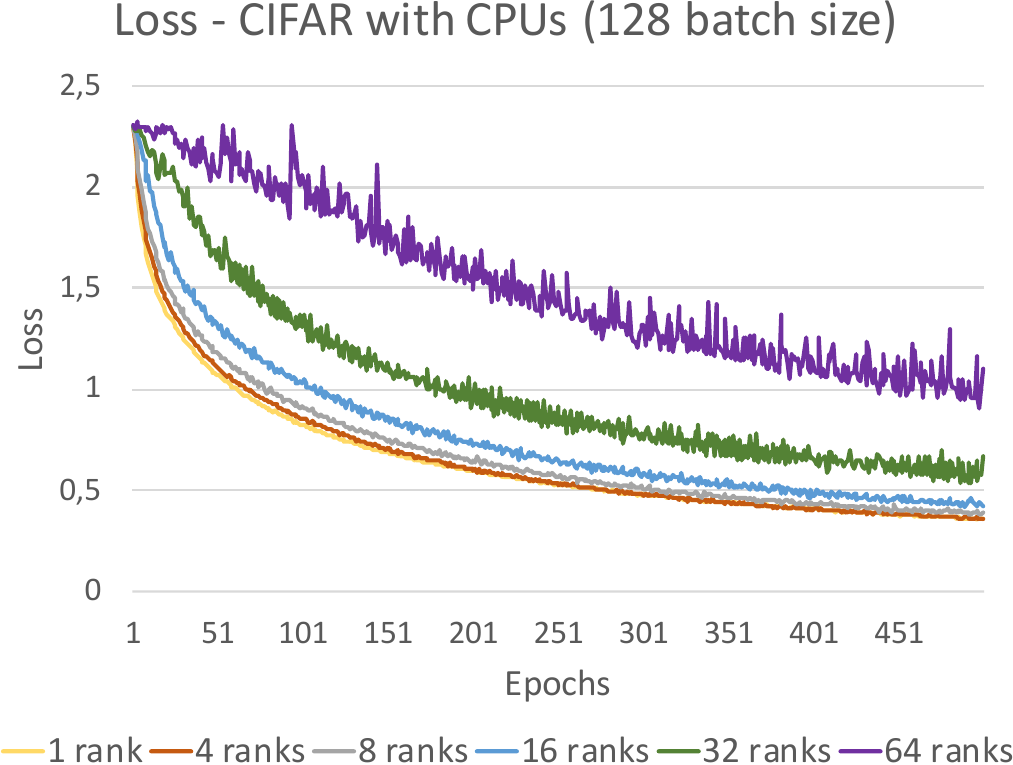}
        \caption{}
        \label{fig:Loss-CIFAR-CPU}
    \end{subfigure}%
    \hspace{0.07\textwidth}
    \begin{subfigure}{0.38\textwidth}
        \centering
        \includegraphics[width=1\linewidth]{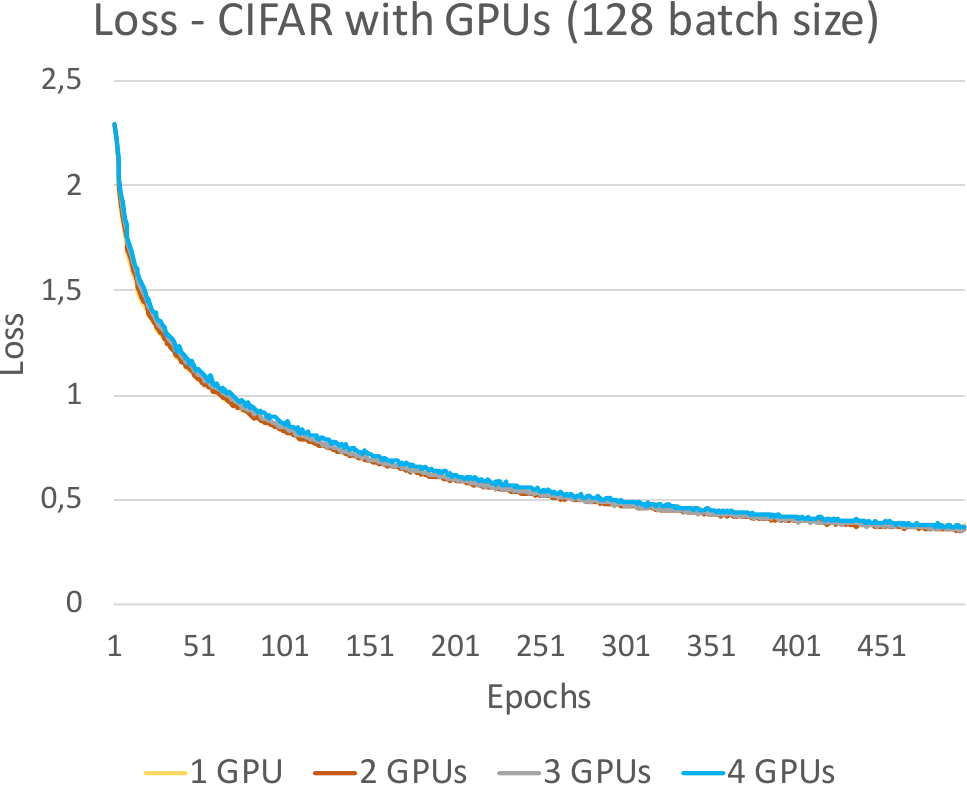}
        \caption{}
        \label{fig:Loss-CIFAR-GPU}
    \end{subfigure}%
    \caption{Convergence analysis of TensAIR on the Word2Vec and CIFAR-10 use-cases.}
    \label{fig:Loss}
\end{figure*}

We first explored TensAIR's ability to converge by determining if and how DASGD might degrade the quality of the trained model (Figure \ref{fig:Loss}). We compared the training loss curve of Word2Vec and CIFAR-10 by distributing TensAIR models from 1 to 4 GPUs using 1 TensAIR rank per GPU (Figures \ref{fig:Loss-W2V-GPU} \& \ref{fig:Loss-CIFAR-GPU}). We additionally explored the models convergence when trained with distributed CPU nodes (Figures \ref{fig:Loss-W2V-CPU} \& \ref{fig:Loss-CIFAR-CPU}). In this second scenario, we trained up to 64 ranks on 16 nodes simultaneously without GPUs. Note that, when using a single TensAIR rank, TensAIR's gradient updates behave as in a synchronous SGD implementation.   

The {\em extremely low variance} among all loss curves shown in Figures \ref{fig:Loss-W2V-CPU} and \ref{fig:Loss-W2V-GPU} demonstrates that our asynchronous and distributed SGD updates do not at all negatively affect the convergence of the Word2Vec models. We assume that this is due to (1) the sparseness of Word2Vec, and (2) a low staleness of the gradients (which are relatively inexpensive to compute and apply for Word2Vec). The low staleness indicates a fast exchange of gradients among models.

In Figure \ref{fig:Loss-CIFAR-CPU}, we however observe a remarkable degradation of the loss when distributing CIFAR-10 across multiple nodes. This is due to the fixed learning rate used on all settings being the same. When distributing dense models on multiple ranks without adapting the mini-batch size, it is well known to result in a degradation of the loss curve (even on synchronous settings). This degradation occurs because the behaviour of training $N$ models with mini-batches of size $b$ is similar to training 1 model with mini-batches of size $N \cdot b$. To mitigate this issue, Horovod increases the learning rate $\alpha$ by the number of ranks used to distribute the model \cite{horovod_documentation}, i.e., $\alpha_{new} = \alpha \cdot N$. Accordingly, in Figure \ref{fig:Loss-CIFAR-GPU}, we again do not see any degradation of the loss when distributing CIFAR-10 because we use a maximum of 4 GPUs. 

\subsection{Speed-up Analysis}\label{sec:speedup}
\begin{figure*}[ht]
    \centering
    \begin{subfigure}{.42\textwidth}
        \centering
        \includegraphics[width=1\linewidth]{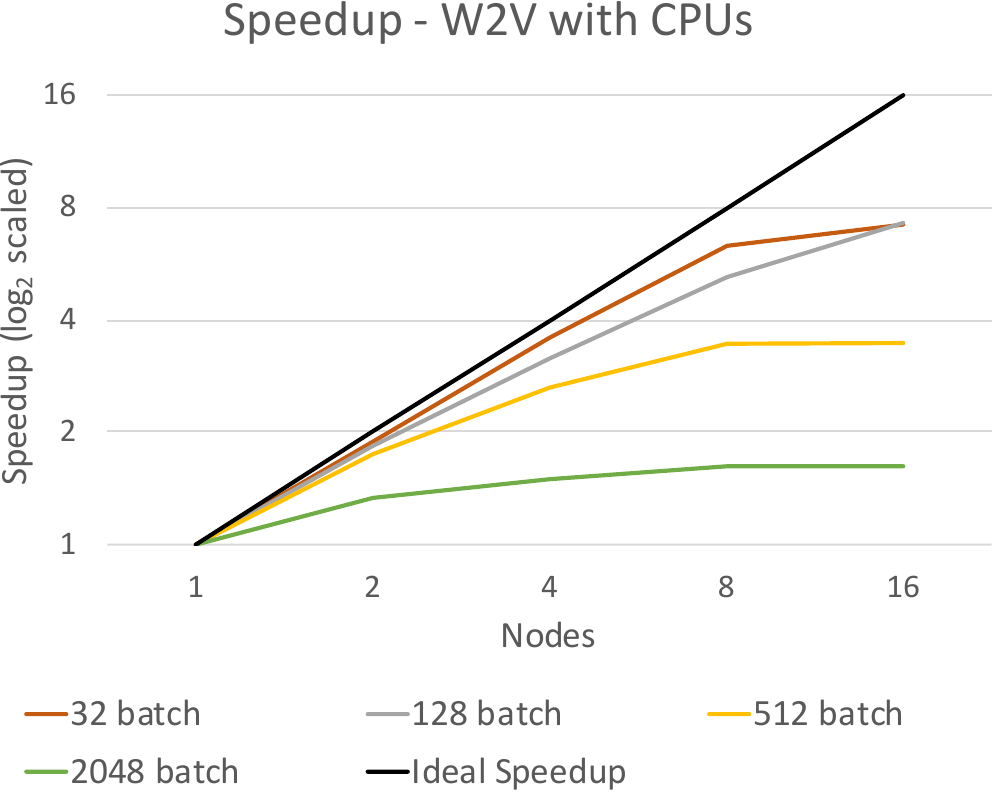}
        \caption{}
        \label{fig:Speedup-W2V-CPU}
    \end{subfigure}%
    \hspace{.07\textwidth}
    \begin{subfigure}{0.42\textwidth}
        \centering
        \includegraphics[width=1\linewidth]{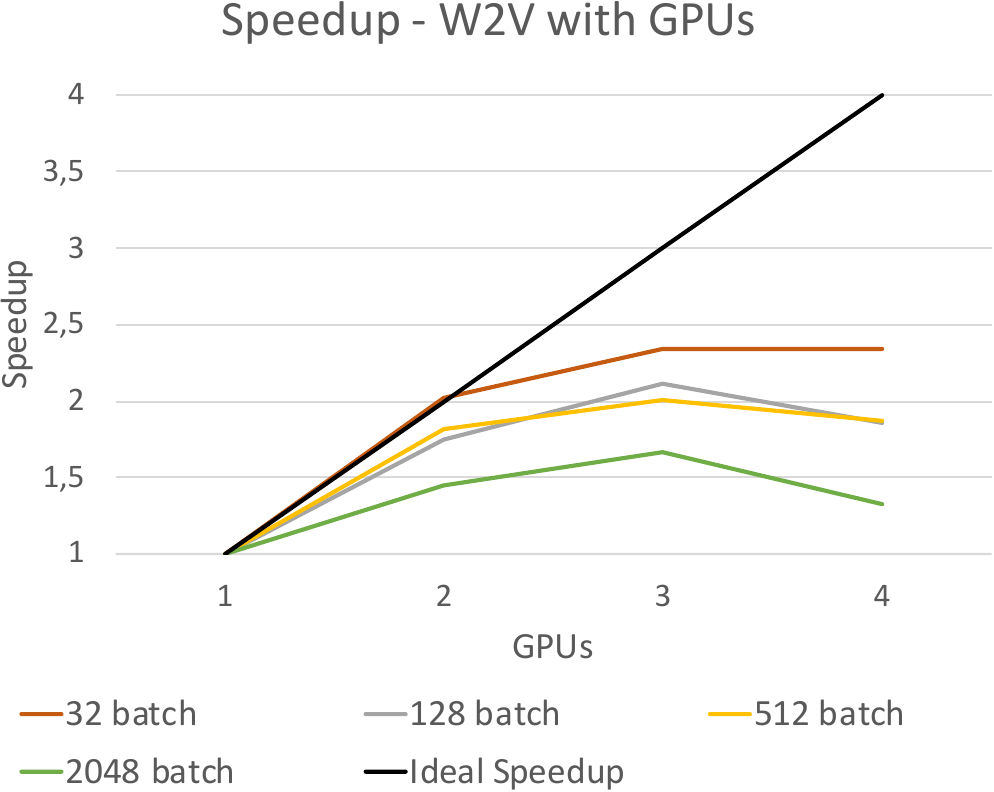}
        \caption{}
        \label{fig:Speedup-W2V-GPU}
    \end{subfigure}%
    \vspace{1em}
    \begin{subfigure}{0.42\textwidth}
        \centering
        \includegraphics[width=1\linewidth]{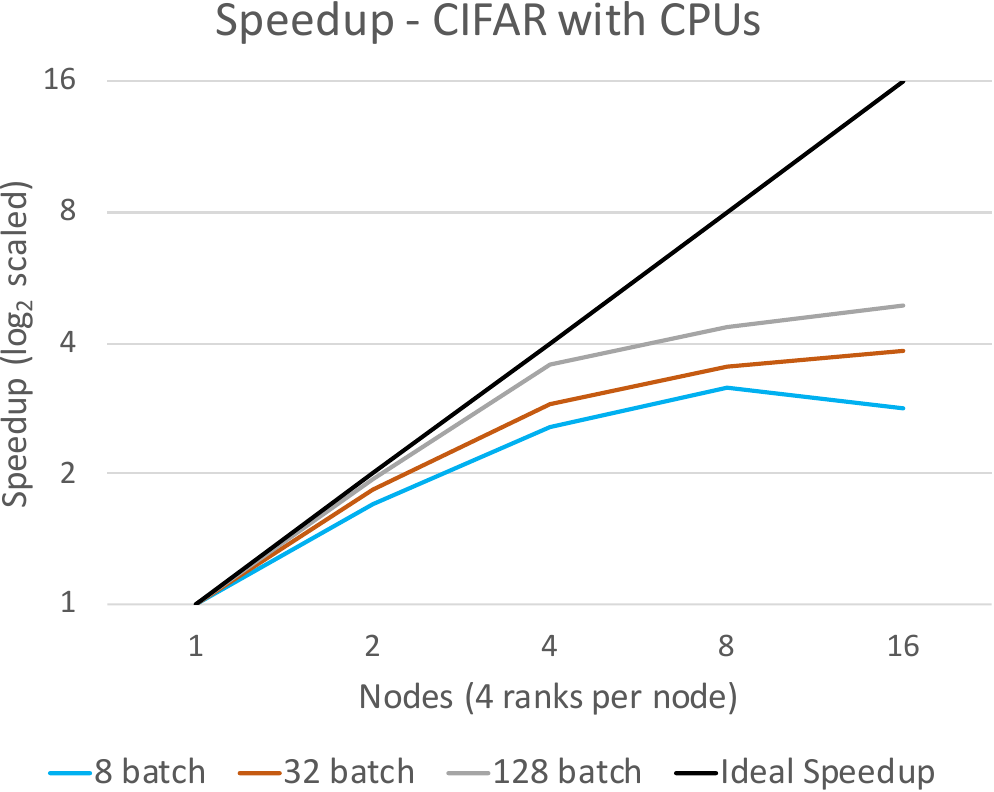}
        \caption{}
        \label{fig:Speedup-CIFAR-CPU}
    \end{subfigure}%
    \hspace{.07\textwidth}
    \begin{subfigure}{0.42\textwidth}
        \centering
        \includegraphics[width=1\linewidth]{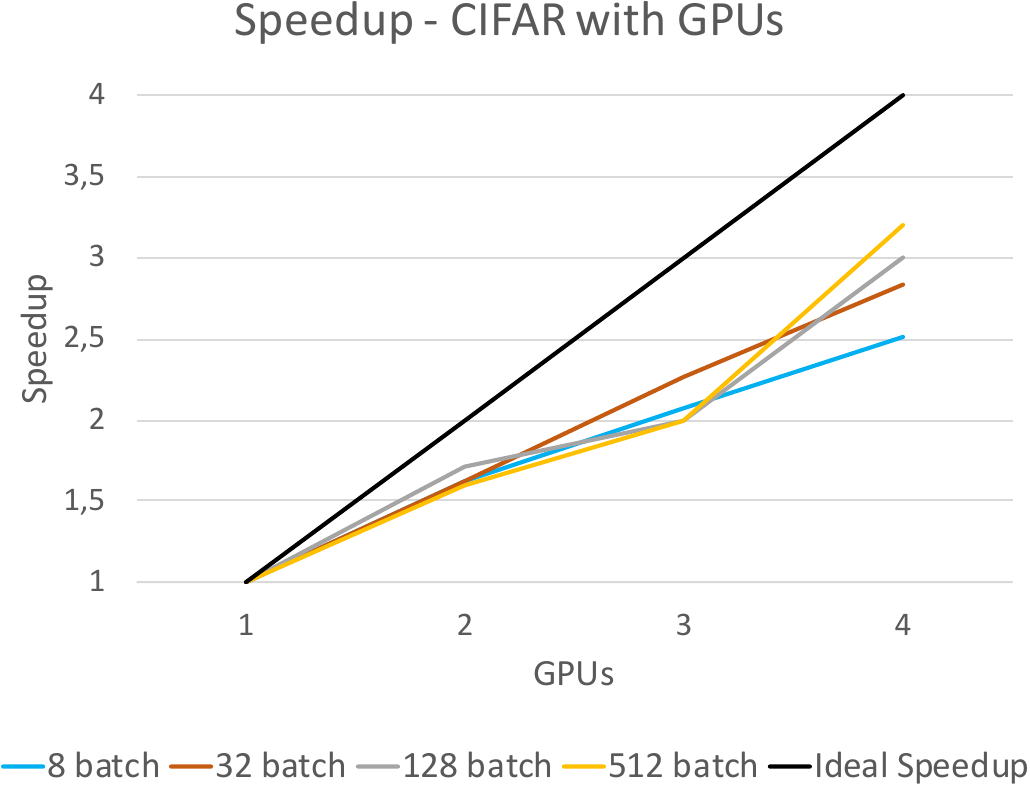}
        \caption{}
        \label{fig:Speedup-CIFAR-GPU}
    \end{subfigure}%
    \caption{Speedup analysis of TensAIR on the Word2Vec and CIFAR-10 use-cases.}
    \label{fig:Speedup}
\end{figure*}

Next, we explore the performance of TensAIR under increasing levels of distribution and with respect to varying mini-batch sizes over both Word2Vec and CIFAR-10. This experiment is also deployed on up to 64 ranks (16 nodes) and up to 4 GPUs (1 node). We observe in Figure \ref{fig:Speedup} that TensAIR achieves a {\em nearly-linear scale-out} under most of our settings. 

In most cases, TensAIR achieves a better speedup when training with smaller mini-batches. This difference is because, differently than the gradient calculation, the gradient application is not distributed and, with smaller mini-batches, more gradients are applied per epoch. Thus, models with expensive gradient computations will have a better scale-out performance. Nevertheless, when gradient calculation is not the bottleneck of the dataflow, one can reduce the computational impact of the gradients application and the network impact of their broadcasts by simply increasing {\em maxBuffer}. For instance, by increasing {\em maxBuffer} in $n$ times, the network complexity and the computational impact of the gradients applications are also expected to be reduce in $n$ times.

\subsection{Baseline Comparison}
\label{sec:baseline}
\begin{figure*}[ht]
    \centering
    \begin{subfigure}{.5\textwidth}
        \centering
        \includegraphics[width=1\linewidth]{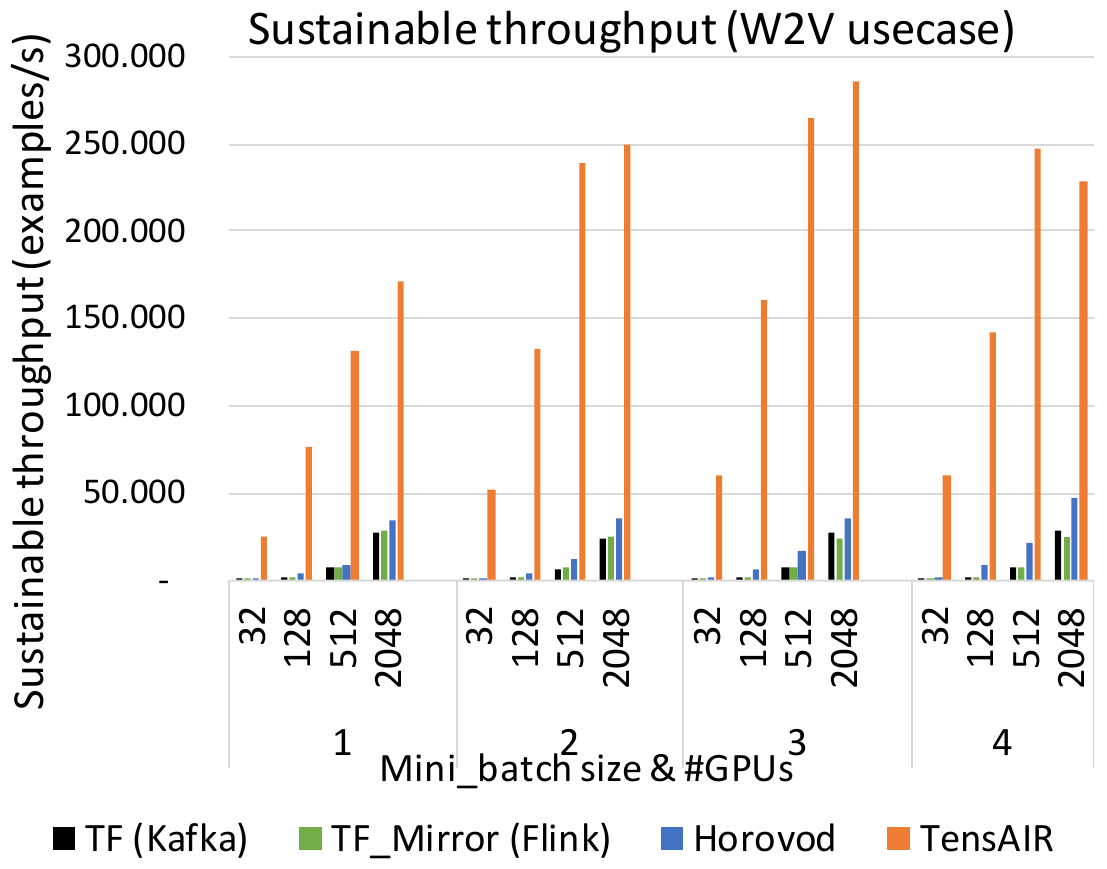}
        \caption{W2V use-case.}
        \label{fig:TensAIRxHOROVOD-W2V}
    \end{subfigure}%
    \begin{subfigure}{0.5\textwidth}
        \centering
        \includegraphics[width=1\linewidth]{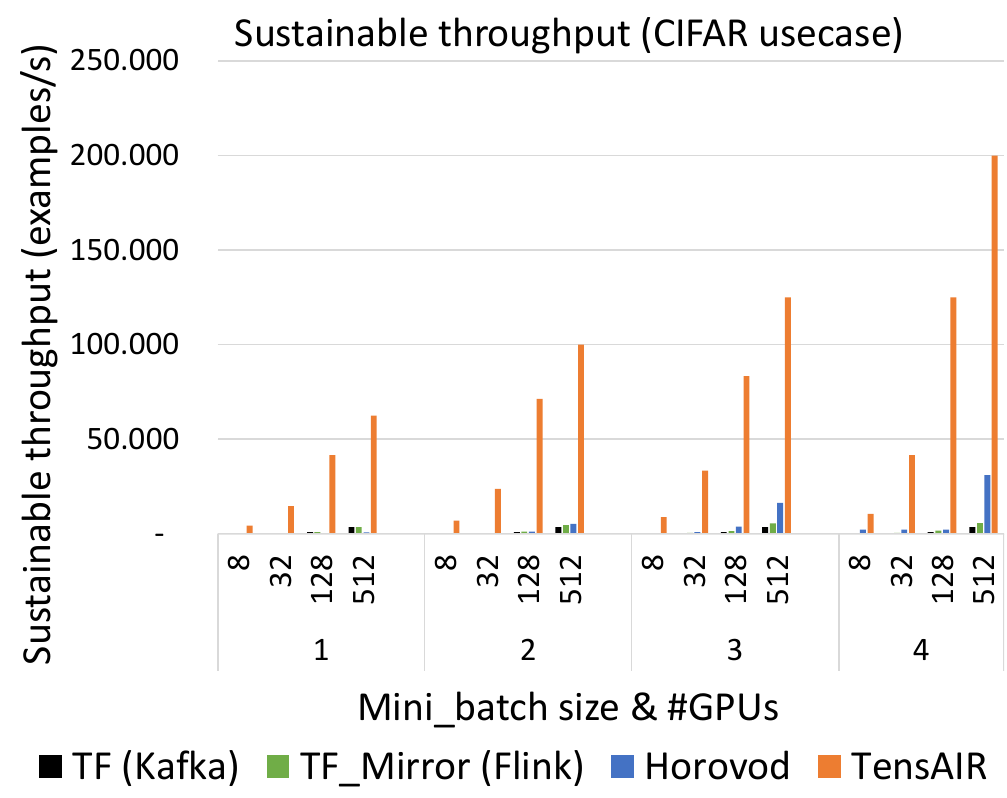}
        \caption{CIFAR-10 use-case.}
        \label{fig:TensAIRxHorovod-CIFAR}
    \end{subfigure}%
    \caption{Throughput comparison between TensAIR, TensorFlow (standard and distributed), and Horovod.}
    \label{fig:TensAIRxHorovod}
\end{figure*}

Apart from TensAIR, it is also possible to train ANNs by using Apache Kafka and Flink as message brokers to generate data streams of varying throughputs. Kafka is already included in the standard TensorFlow I/O library ({\tt tensorflow\_io}), which however allows no actual distribution in the training phase \cite{kafka_tensorflow}. Flink, on the other hand, employs the distributed TensorFlow API ({\tt tensorflow.dis\-tribute}). However, we were not able to run the provided {\em dl-on-flink} use-case \cite{dl-on-flink} even after various attempts on our HPC setup. We therefore report the direct deployment of our Word2Vec and CIFAR-10 use-cases (Figures \ref{fig:TensAIRxHOROVOD-W2V} \& \ref{fig:TensAIRxHorovod-CIFAR}) on both the standard and distributed TensorFlow APIs (the latter using the {\tt MirroredStrategy} option of {\tt tensorflow.distribute}). We thereby, simulate a streaming scenario by feeding one mini-batch per training iteration into TensorFlow, which yields a very optimistic upper-bound for the maximum throughput that Kafka and Flink could achieve. In a similar manner, we also determined the maximum throughput of Horovod \cite{sergeev2018horovod}, which is however not a streaming engine by default.

In Figures \ref{fig:TensAIRxHOROVOD-W2V} and \ref{fig:TensAIRxHorovod-CIFAR}, we see that TensAIR clearly surpasses both the standard and distributed TensorFlow setups as well as Horovod. This occurs because, as opposed to TensAIR, their architectures were not developed to train on batches arriving from data streams. Thus, in a streaming scenario, the overhead of transferring the training data to the worker nodes increases by the number of training steps. On the other hand, TensAIR was designed to train ANN models from high throughput data streams in real-time. Thus, the transfer of training data overhead is mitigated by the asynchronous protocol adopted and the training is speed-up by DASGD. This allows TensAIR to (1) reduce both computational resources and idle times while the data is being transferred, and (2) have an optimized buffer management for incoming mini-batches and outgoing gradients, respectively.

In our experiments, we could sustain a maximum training rate of 285,560 training examples per second on Word2Vec and 200,000 images per second on CIFAR-10, which corresponds to sustainable throughputs of 14.16 MB/s and 585 MB/s respectively. We reached these values by training with 3 GPUs on Word2Vec and 4 GPUs on CIFAR-10. Note that, while using more than 3 GPUs simultaneously, TensAIR did not achieve better sustainable throughput in the W2V usecase due to the relatively low complexity of the gradient calculations. In this scenario, the training bottleneck, typically associated with gradient calculation, shifted to the gradient application when using more than 3 GPUs, as the former is not distributed. Nevertheless, this issue can be mitigated by simply increasing the variable $maxBuffer$ (as explained in Section \ref{sec:speedup}). This adjustment, delays the communication among distributed models while reducing the locally applied gradients by a factor of $maxBuffer$. 

\subsection{Sentiment Analysis of COVID19} \label{sec:twitter}

Here, we exemplify the benefits of training an ANN in real-time from streaming data. To this end, we analyze the impact of concept drifts on a sentiment analysis setting, specifically drifts that occurred during and due to the COVID19 pandemic. First, we trained a large Word2Vec model using 20\% of English Wikipedia plus the Sentiment140 dataset \cite{go2009twitter}. Then, we trained an LSTM model \cite{tensorflow_tutorial_lstm} using the Sentiment140 dataset together with the word embeddings we trained previously. After three epochs, we reached 78\% accuracy on the training and the test set. However, language is always evolving. Thus, this model may not sustain its accuracy for long if deployed to analyze streaming data in real-time. We exemplify this by fine-tuning the word embeddings with 2M additional tweets published from November 1st, 2019 to October 10th, 2021 containing the following keywords: \textit{covid19}, \textit{corona}, \textit{coronavirus}, \textit{pandemic}, \textit{quarantine}, \textit{lockdown}, \textit{sarscov2}. Then, we compared the previously trained word embeddings and the fine-tuned ones and found an average cosine difference of only 2\%. However, despite being small, this difference is concentrated onto specific keywords.

\begin{table}[h]
\centering
\begin{tabular}{l|lllll}
\textbf{Term}       & rt    & corona & pandemic & booster & 2021  \\ \hline
\textbf{Difference} & 0.728 & 0.658  & 0.646    & 0.625   & 0.620
\end{tabular}
\caption{Cosine differences after updating word embeddings.}
\label{tab:words}
\end{table}

As shown in Table \ref{tab:words}, keywords related to the COVID-19 pandemic are the ones that most suffered from a concept drift. Take as example \textit{pandemic}, \textit{booster} and \textit{corona}, which had over 62\% of cosine difference before and after the Word2Vec models have been updated.
Due to the concept drift, the sentiment over specific terms and, consequently, entire tweets also changed. One observes this change by comparing the output of our LSTM model when: (1) inputting tweets embedded with the pre-trained word embeddings; (2) inputting tweets embedded with the fine-tuned word embeddings. Take as an example the sentence ``\textit{I got corona.}'', which had a sentiment of $+2.0463$ when predicted with the pre-trained embeddings; and $-2.4873$ when predicted with the fine-tuned embeddings. Considering that the higher the sentiment value the more positive the tweet is, we can observe that {\em corona} (also representing a brand of a beer) was seen as positive and now is related to a very negative sentiment.

\begin{figure*}[ht]
  \centering
  \includegraphics[width=0.78\textwidth]{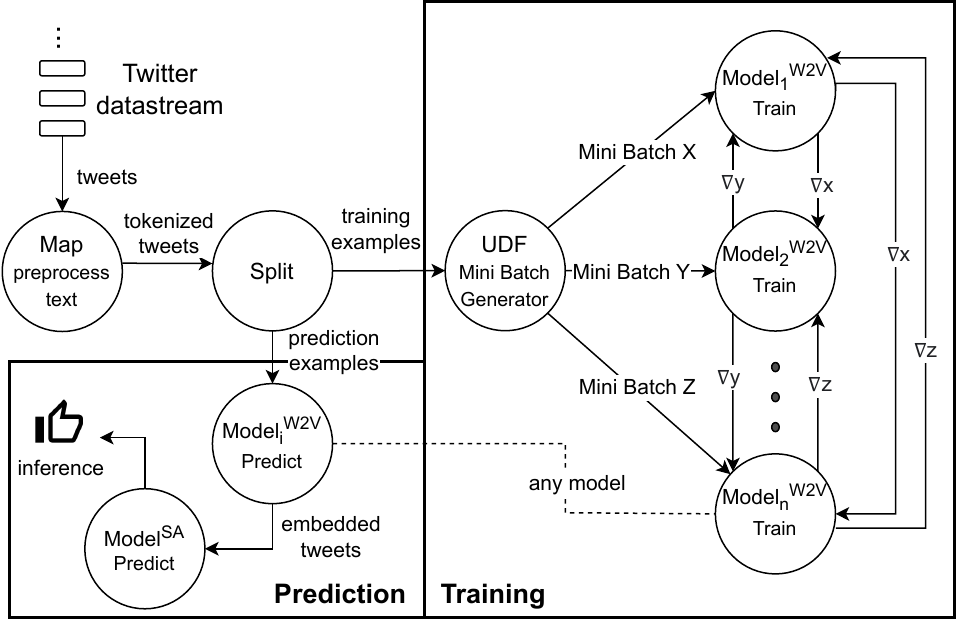}
  \caption{TensAIR dataflow with $n$ distributed {\tt Model}$^{W2V}$ instances and a single instance of {\tt  Map}, {\tt Split}, {\tt UDF} and {\tt Model}$^{SA}$.}
  \label{fig:dataflow2}
\end{figure*}

To tackle concept drifts in this use-case, we argue that TensAIR with its OL components (as depicted in Figure \ref{fig:dataflow2}) could be readily deployed. A real-time pipeline with Twitter would allow us to constantly update the word embeddings (our sustainable throughput would be more than sufficient compared to the estimated throughput of Twitter). Consequently, the sentiment analysis algorithm would always be up-to-date with respect to such concept drifts.

Figure~\ref{fig:dataflow2} depicts the dataflow for a {\em Sentiment Analysis} (SA) use-case on a Twitter data stream. This dataflow predicts the sentiments of live tweets using a pre-trained ANN model ({\tt Model}$^{SA}$). However, it does not rely on pre-defined word embeddings. The dataflow constantly improves its embeddings on a second {\em Word2Vec} (W2V) ANN model ({\tt Model}$^{W2V}$), which it trains using the same input stream as used for the predictions. By following a passive concept-drift adaptation strategy, it can adapt its sentiment predictions in real-time based on changing word distributions among the input tweets. Moreover, it does not require any sentiment labels for newly streamed tweets at {\tt Model}$^{SA}$, since only {\tt Model}$^{W2V}$ is re-trained in a self-supervised manner by generating mini-batches of word pairs $(x,y)$ directly from the input tweets.

Our SA dataflow starts with {\tt Map} which receives tweets from a Twitter input stream (implemented via cURL or a file interface) and tokenizes the tweets based on the same word dictionary also used by {\tt Model}$^{W2V}$ and {\tt Model}$^{SA}$. \texttt{Split} then identifies whether the tokenized tweets shall be used for re-training the word embeddings, for sentiment prediction, or for both. If the tokenized tweets are selected for training, they are turned into mini-batches via the {\tt UDF} operator. The $(x,y)$ word pairs in each mini-batch $X$ are sharded across {\tt Model}$^{W2V}_1$, $\ldots$, {\tt Model}$^{W2V}_n$ with a standard hash-partitioner using words $x$ as keys. {\tt Model}$^{W2V}$ implements a default skip-gram model. If the tokenized tweets are selected for prediction, a tweet is vectorized by using the word embeddings obtained from any of the {\tt Model}$^{W2V}$ instances and sent to the pre-trained {\tt Model}$^{SA}$ which then predicts the tweets' sentiments. 

\section{Conclusions}
\label{sec:conclusion}

OL is an emerging area of research which still has not extensively explored the real-time training of ANNs. In this paper, we introduced TensAIR, a novel system for real-time training of ANNs  from data streams. It uses the asynchronous iterative routing (AIR) protocol to train and predict ANNs in a decentralized manner. The two main features of TensAIR are: (1) leveraging the iterative nature of SGD by updating the ANN model with fresh samples from the data stream instead of relying on buffered or pre-defined datasets; (2) its fully asynchronous and decentralized architecture used to update the ANN models using decentralized and asynchronous SGD (DASGD). Due to those two features, TensAIR achieves a nearly linear scale-out performance in terms of sustainable throughput and with respect to its number of worker nodes. Moreover, it was implemented using TensorFlow, which facilitates the deployment of diverse use-cases. Therefore, we highlight the following capabilities of TensAIR: (1) processing multiple data streams simultaneously; (2) training models using either CPUs, GPUs, or both; (3) training ANNs in an asynchronous and distributed manner; and (4) incorporating user-defined dataflow pipelines. We empirically demonstrate that---in a real-time streaming scenario---TensAIR supports from 4 to 120 more sustainable throughput than Horovod and both the standard and distributed TensorFlow APIs (representing upper bounds for Apache Kafka and Flink extensions).

As future work, we believe that TensAIR may also lead to novel online learning use cases which were previously considered too complex but now become feasible due to the very good sustainable throughput of TensAIR. Specifically, we intend to study similar learning tasks over audio/video streams, which we see as the main target domain for stream processing and OL. To reduce the computational cost of training an ANN indefinitely, we shall also investigate how different active concept-drift detection algorithms behave under an OL setting with ANNs.

\begin{acks}
This work is funded by the Luxembourg National Research Fund under the PRIDE program (PRIDE17/12252781). The paper benefited from helpful comments and suggestions by Ovidiu Cristian Marcu. The experiments presented in this paper were carried out
using the HPC facilities of the University of Luxembourg~\cite{VCPKVO_HPCCT22}
(see \texttt{\href{http://hpc.uni.lu}{hpc.uni.lu}}).
\end{acks}

\bibliographystyle{ACM-Reference-Format}
\bibliography{sample-base}

\end{document}